\pdfoutput=1

\documentclass[11pt]{article}

\usepackage[]{acl}
\usepackage{changepage}
\usepackage{tcolorbox}
\usepackage{fontawesome}
\usepackage{booktabs}
\usepackage{adjustbox}

\usepackage{alltt}
\usepackage{array}  

\newcolumntype{M}[1]{>{\centering\arraybackslash}m{#1}}

\usepackage{graphicx}
\usepackage{amsmath}
\usepackage{float}
\usepackage{amssymb}
\usepackage{booktabs}
\usepackage{enumitem}

\usepackage[]{hyperref}

\usepackage{tcolorbox}
\tcbuselibrary{breakable,xparse,skins}
\usepackage{listings}
\usepackage{subcaption}
\usepackage{arydshln}

\definecolor{codegreen}{HTML}{006600}
\definecolor{codegray}{rgb}{0.5,0.5,0.5}
\definecolor{codepurple}{rgb}{0.58,0,0.82}
\definecolor{backcolour}{HTML}{f0f5f5}
\definecolor{agentcolor}{HTML}{993399}
\definecolor{clientcolor}{HTML}{333333}
\definecolor{referencecolor}{HTML}{1174ad}
\definecolor{rougecolor}{HTML}{B85450}
\definecolor{textbgcolor}{HTML}{F5F5F5}
\definecolor{promptbgcolor}{HTML}{ecf9ec}
\definecolor{workflowbgcolor}{HTML}{e6ecff}
\definecolor{slotcolor}{HTML}{cc7a00}
\definecolor{specialtokencolor}{HTML}{b30000}

\usepackage[capitalize]{cleveref}
\crefname{section}{Sec.}{Secs.}
\Crefname{section}{Section}{Sections}
\Crefname{table}{Table}{Tables}
\crefname{table}{Tab.}{Tabs.}

\begin{document}

\title{MAGID: An Automated Pipeline for Generating Synthetic Multi-modal Datasets}

\author{Hossein Aboutalebi\textsuperscript{\thanks{\hspace{0.2cm}Work conducted while interning at AWS AI Labs.}\hspace{0.1cm}, \faFlag} \hspace{.5em}  Hwanjun Song\textsuperscript{\faAmazon} \hspace{.5em} \textbf{Yusheng Xie}\textsuperscript{\faAmazon} \hspace{.5em} \textbf{Arshit Gupta}\textsuperscript{\faAmazon}\\ \hspace{.5em} \textbf{Justin Sun}\textsuperscript{\faAmazon} \hspace{.5em} \textbf{Hang Su}\textsuperscript{\faAmazon}\hspace{.5em} \textbf{Igor Shalyminov}\textsuperscript{\faAmazon} \hspace{.5em} \textbf{Nikolaos Pappas}\textsuperscript{\faAmazon} \hspace{.5em} \\ \textbf{Siffi Singh}\textsuperscript{\faAmazon} \hspace{.5em}  \textbf{Saab Mansour}\textsuperscript{\faAmazon}\\
       \textsuperscript{\faFlag} Cheriton School of Computer Science, University of Waterloo\\
       \textsuperscript{\faAmazon} AWS AI Labs\\
        \texttt{haboutal@uwaterloo.ca}}
\maketitle

\begin{abstract}

Development of multimodal interactive systems is hindered by the lack of rich, multimodal (text, images) conversational data, which is needed in large quantities for LLMs. Previous approaches augment textual dialogues with retrieved images, posing privacy, diversity, and quality constraints. In this work, we introduce \textbf{M}ultimodal \textbf{A}ugmented \textbf{G}enerative \textbf{I}mages \textbf{D}ialogues (MAGID), a framework to augment text-only dialogues with diverse and high-quality images \footnote{code link: https://github.com/amazon-science/MAGID}. Subsequently, a diffusion model is applied to craft corresponding images, ensuring alignment with the identified text. Finally, MAGID incorporates an innovative feedback loop between an image description generation module (textual LLM) and image quality modules (addressing aesthetics, image-text matching, and safety), that work in tandem to generate high-quality and multi-modal dialogues. We compare MAGID to other SOTA baselines on three dialogue datasets, using automated and human evaluation. Our results show that MAGID is comparable to or better than baselines, with significant improvements in human evaluation, especially against retrieval baselines where the image database is small. 

\end{abstract}


\section{Introduction}
\label{sec:intro}

In recent years, advancements in large language models\,(LLMs) have expanded possibilities and research directions in AI, with studies highlighting their extensive capabilities in handling dialogue datasets~\cite{liu2023summary, penedo2023refinedweb}. 
Specifically, there is a growing interest in their application to multi-modal dialogue datasets, given that \emph{sharing images} is an integral aspect of human-human conversations~\cite{alayrac2022flamingo, openai2023gpt, liu2023visual}.

Several multi-modal dialogue datasets like MMDialog~\cite{feng2022mmdialog}, DialogCC~\cite{lee2022dialogcc}\footnote{A recently released version of DialogCC utilizes LLM \cite{lee2023dialogcc}. At the time of writing this paper, we did not have access to the newer version.}, and PhotoChat~\cite{zang2021photochat} have been introduced for training multi-modal LLMs. These datasets either use a retrieval-based approach, pulling images from set image banks, such as MS-COCO \cite{lin2014microsoft}, or restrict the dialogue to only one image per conversation, even if they involve real human-human chats. Moreover, when leveraging real-world datasets from platforms like social media, issues related to privacy concerns and image quality become significant challenges for training.

\begin{figure*}[ht]
\includegraphics[width=16.2cm]{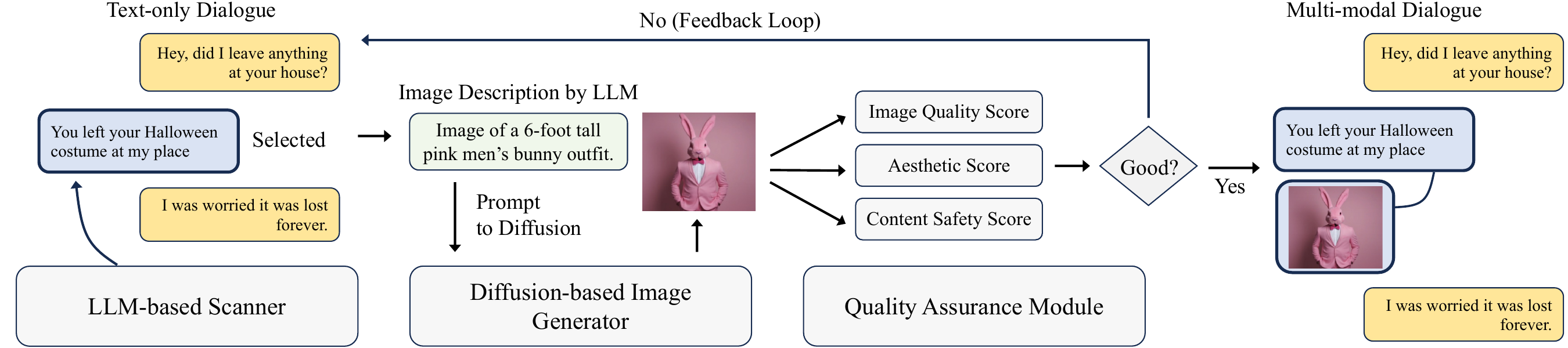}
\caption{Overview of the MAGID framework. MAGID consists of three components: (1) LLM-based scanner to identify suitable utterances to augment with images, (2) diffusion-based image generator to create realistic images, and (3) quality assurance module to enhance the image quality, aesthetic and safety scores. The text-only dialogue is automatically converted to multi-modal dialogue using MAGID.}
\label{fig:magi_f}
\vspace*{-0.3cm}
\end{figure*}

As a result, these methods limit the diversity of images since the small image database cannot adequately capture the wide range of real human-human conversations \cite{lee2021constructing, lee2022dialogcc}. Additionally, they face challenges stemming from low-quality images containing harmful and private content \cite{feng2022mmdialog} and shortage of accessible data \cite{lee2022dialogcc}, particularly when utilizing real human-human conversations from social media sources.

To address these challenges, we propose \textbf{MAGID}, a \emph{generative}-based multi-modal dialogue creation framework. As illustrated in Figure \ref{fig:magi_f}, MAGID aims at converting existing text-only data into context-enriched multi-modal data by addressing the two research challenges: {(i) how to find the most suitable utterances that can be enhanced by adding images} and {(ii) how to generate realistic and diverse images that do not have harmful and private contents}. 

In the former case, we introduce an \emph{LLM-based scanner} designed to pinpoint utterances requiring images and subsequently generate corresponding image descriptions, leveraging chain-of-thought prompting. In the latter case, we employ a \emph{diffusion-based image generator}, adept at crafting images with notable diversity, drawing upon the generated image descriptions as its input. Additionally, a \emph{quality assurance} module is incorporated into our framework to ensure both the congruence and the quality of the produced images, thereby preserving coherence and fidelity within the multi-modal dialogue. Should the generated image not satisfy the criteria of this module, MAGID initiates a feedback loop, revisiting the processes of prompt and image generation.

Distinct from numerous previous endeavors that have depended on image-retrieval techniques for curating multi-modal datasets \cite{lee2021constructing, lee2022dialogcc}—\textbf{a method that might result in restricted image diversity and potential mismatch with the dialogue existing utterances}—we employ the generative model Stable Diffusion XL~\cite{podell2023sdxl}. By training on billions of images \cite{schuhmann2022laion}, this approach guarantees an output that is both rich and varied. Such outputs align well with the conversational context provided by the LLM feedback, thereby elevating the quality and diversity of our multi-modal dataset.

Our framework aligns with prior studies using text-only datasets\,\cite{lee2021constructing, lee2022dialogcc}, but it addresses the limitations associated with their retrieval-based strategies by employing a generative-based data creation method. Unlike \citet{liu2023visual, lee2021constructing}, we do not restrict the inclusion of only one image per dialogue. Consequently, MAGID generates synthetic yet more realistic multi-modal dialogue datasets
thus mitigating data accessibility issues and facilitating the development of advanced multi-modal models.

To summarize, our main contributions are:
\begin{itemize}[leftmargin=12pt]
\item We present MAGID, a generative-based multi-modal dialogue data creation framework that addresses the limitation of retrieval-based approaches.
\vspace*{-0.15cm}
\item We conduct experiments using various prompt engineering strategies to optimize interactions between the LLM-based scanner and the diffusion-based image generator.
\vspace*{-0.15cm}
\item We propose a novel quality assurance design to  control the performance of generative models effectively. 
\vspace*{-0.15cm}
\item We provide a medium-sized dataset as a proof of concept to showcase the effectiveness of MAGID pipeline (section \ref{dataset_poc}). 
\vspace*{-0.15cm}
\item We conduct extensive human evaluations on the dataset and test multiple LLM models to ensure robustness and reliability.
\end{itemize}

\begin{figure*}[]
    \centering
    \begin{tcolorbox}[colback={promptbgcolor},title={\small Zero shot prompt},enhanced,attach boxed title to top right={yshift=-3.5mm, xshift=-5mm}, colbacktitle=white, coltitle=black, top=12pt]  
        \tiny
You are an AI assistant that helps augment textual dialogues with engaging images. As input, you will receive a conversation between people which is represented as a sequence of utterances. As output, you will generate a description of images that can support the utterances in the conversation. 
    
    The format of the input is \textcolor{specialtokencolor}{'Utterance i: ...'} where 
    \textcolor{specialtokencolor}{'i'} denotes the order of the Utterance in the conversation. Given this query, you output in the format of
    
    \textcolor{slotcolor}{
    \textless result\textgreater Utterance i: image\_description\textless/result\textgreater ~\textless reason\textgreater explanation\_of\_choice \textless/reason\textgreater} 
    
    where \textcolor{specialtokencolor}{'i'} is the Utterance 
    in the conversation and \textcolor{specialtokencolor}{'image\_description'} is the  short text description of an image that can be followed by that Utterance 
    that can make the conversation more engaging. You should only identify the most appropriate utterances in the conversation. 
    
    The text inside \textcolor{slotcolor}{\textless reason\textgreater explanation\_of\_choice\textless/reason\textgreater} is the explanation of why you picked the utterance with the image description.
   \end{tcolorbox}
    \vspace*{-0.45cm}
   \caption{The zero-shot prompt of the scanner module (Section~\ref{sec:scanner}) which selects turns in the dialogue to augment with images and generates descriptions of those images. Additional few-shot and chain-of-thought prompts are provided in the supplementary materials (section \ref{COT_prompt}).  }
    \vspace*{-0.45cm}
   \label{fig:zero_shot}
\end{figure*}

\section{Related Works}

\subsection{Generative Models}
Recent advances in Generative AI has started new trends in expanding capabilities of existing deep learning models. In NLP, works like \cite{radford2019language, ouyang2022training} have shown importance of training data to build better LLM models. In this regard, recent LLM models like Falcon-40b-Instruct \cite{penedo2023refinedweb}, Koala 13b \cite{geng2023koala}, LLaMA 13b \cite{touvron2023llama}, OpenLLaMA \cite{touvron2023llama}, and Vicuna 13b \cite{vicuna2023} use better curated training datasets to achieve higher performances. In this regard, paper like \citet{christiano2017deep} has shown the dramatic impact of using higher quality data (from human feedback) in faster training. Yet, using human feedback and crowd-sourcing is not always cheap. To address this, emerging works like \cite{veselovsky2023artificial, kamalloo2023hagrid} suggests that LLM has the capabilities of performing the task of human generated dataset. In addition, diffusion models in computer vision have shown promising results in generating images indistinguishable from real ones \cite{podell2023sdxl, ho2020denoising}. 
Finally, recent works focus on building multi-modal LLM models including GPT-4 \cite{openai2023gpt}, LLaVA \cite{liu2023llava}, AnyMAL\cite{moon2023anymal} which supports any modality. Specifically, LLaVA accepts multi-modal input, combining image and text embeddings to generate text-only output. 

\begin{figure*}[ht]
\centering
\includegraphics[width=1.0\textwidth]{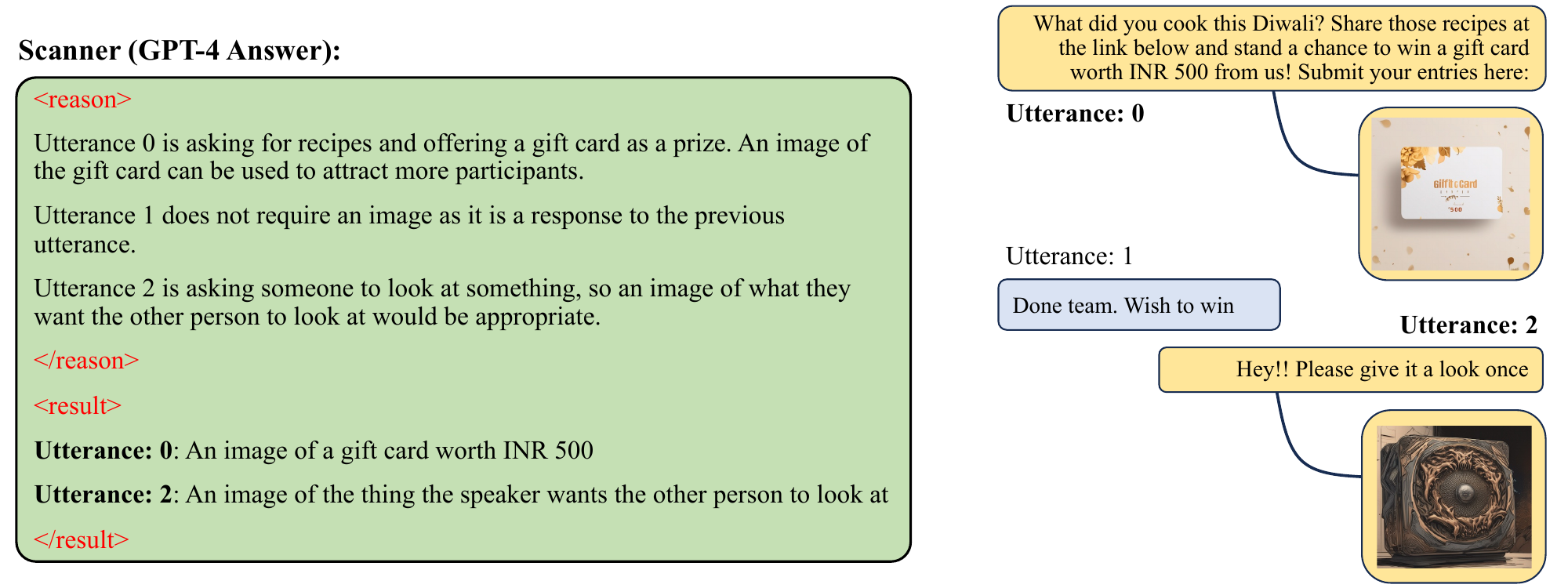}
\vspace*{-0.7cm}
\caption{MAGID's chain of thought prompting facilitates debugging and identification of corner cases, utilizing the SDXL 1.0 diffusion model and GPT-4 \cite{openai2023gpt}. The depicted conversation is sourced from a real human-human interaction in the MMDialog dataset \cite{feng2022mmdialog}.}
\label{fig:xai}
\vspace*{-0.3cm}
\end{figure*}

\subsection{Multi-modal Dataset Creation}

There are also works which focus on generating multi-modality datasets. In particular, MMDD \cite{lee2021constructing} and DialogCC \cite{lee2022dialogcc} use image-retrieval approaches to augment text-only datasets to multi-modal datasets. PhotoChat \cite{zang2021photochat} hires workers to discuss a particular image to build the dataset. MMDialog \cite{feng2022mmdialog} collect multi-modal conversations from internet to build the dataset which can potentially pose privacy concern to use as training set. 
There are also works \cite{wang2023internvid, corona2021meva, corona2020meva, ciliberto2021opportunity++, abdrakhmanova2021speakingfaces} which focuses modality beyond text and image including video and voice. For example, \citet{corona2021meva} provide a dataset that contains videos for activity detection. IntenVid \cite{wang2023internvid} is another example that contains video in addition to text.

\section{MAGID Pipeline}

In transitioning from text-only to multi-modal dialogue, there exist two core challenges. The first is the identification of the most suitable utterances within the dialogue that can be enhanced by images. The second is the creation of corresponding, accurate images that align with the selected utterances. In this regard, we need to ensure a harmonious and coherent match between the image and the text, achieving acceptable image-text alignment.

We have addressed these challenges through the implementation of the following three key modules in Figure \ref{fig:magi_f}, namely LLM-based scanner, diffusion-based image generator, and quality assurance module, which are detailed in the subsequent sections.

\subsection{MAGID Scanner}
\label{sec:scanner}

The primary objective of this module is to identify suitable utterances that can be visually represented by an image. Achieving best performance  requires precise control over the behavior of the LLM model. We use prompt engineering and special formatting to control the output of LLM.

We experimented with three prompt engineering strategies to fine-tune the system prompts of the LLM:

\begin{itemize}[leftmargin=12pt]
\item \textbf{Zero-shot prompting:} The LLM is provided with only the format of the input and the expected output, along with a general problem description. Figure \ref{fig:zero_shot} shows an example of the zero-shot prompt.
\item \textbf{Few-shot example prompting:} Besides the information provided in zero-shot prompting, LLM is also supplied with several input–output exemplars to demonstrate the anticipated response from the LLM model \cite{brown2020language}.
We have included this type of prompt in supplementary materials (section \ref{COT_prompt}).
\item \textbf{Chain of Thought prompting:} As per \cite{wei2022chain}, this prompting strategy involves imparting a series of intermediate reasoning steps for each example, facilitating the LLM model's capacity for more advanced reasoning. 
Please refer to supplementary materials for example of this prompt (section \ref{COT_prompt}). 
\end{itemize}

In section \ref{prompt_exp}, we evaluated these prompting strategies. Based on the findings, we selected Chain of Thought prompting as the optimal choice for our MAGID framework.

\subsection{Controlling LLM Output Format}

We introduce a method that seeks to streamline the structuring of LLMs outputs by employing HTML-like tags, aiming to facilitate easier parsing and to shed light on the decision-making process. The utilization of ${\rm <result>}$ and ${\rm <reason>}$ tags is intended to envelope answers and rationales respectively, potentially making post-processing more straightforward and offering a degree of transparency into the model's reasoning, which may be beneficial for debugging purposes.

Figure \ref{fig:xai} demonstrates the impact of using the proposed HTML formatting  inside chain of thought prompt, revealing how meticulous analysis of responses identifies corner cases and ensures contextual congruency in produced images. Whereas the first image aligns with preceding text, the second lacks context. The ${\rm <reason>}$ tag discloses that phrases like "give it a look" influenced image generation. To enhance contextual relevance and model reliability, the system prompt has been refined to instruct the LLM to only generate images when paired with a detailed description, thereby avoiding contextual discrepancies.

\begin{figure*}[htbp]
\vspace{-1cm}
\includegraphics[width=1.0\textwidth]{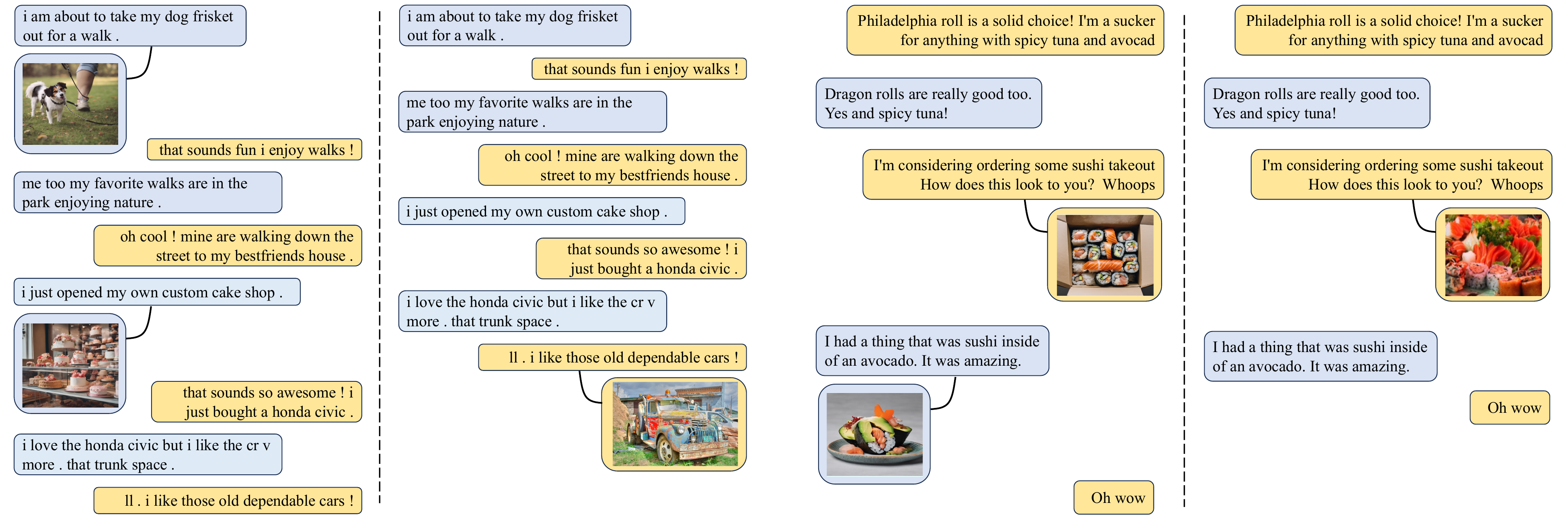}
{\small \hspace*{1.2cm} (a) MAGID (left) vs. MMDD (right). \hspace*{2.35cm} (b) MAGID (left) vs. PhotoChat (right).}
\vspace*{-0.2cm}
\caption{Qualitative comparison of MAGID with an image retrieval-based synthetic MMDD and a real human image-based PhotoChat datasets.}
\label{fig:all_compare}
\vspace*{-0.4cm}
\end{figure*}

\subsection{MAGID Image Generator}

As illustrated in Figure \ref{fig:magi_f}, the LLM model's image prompts are used by the diffusion model to generate corresponding images. In this regard, given the success of diffusion models in superior image generation \cite{rombach2022high, ho2020denoising}, were chosen over GANs \cite{goodfellow2014generative}. Models tested included SDXl 1.0, SDXL 0.9, and Stable Diffusion versions from Stability AI \cite{podell2023sdxl}, with a detailed comparison in supplementary materials (section \ref{diffuse_compare}).

Ultimately, SDXl 1.0 was chosen for its state-of-the-art capabilities, bolstering the quality and reliability of the generated
images of the MAGID dataset. Nevertheless, future model developments can be incorporated to refine our MAGID dataset generation.

\subsection{MAGID Quality Assurance}

The Quality Assurance (QA) module is essential for improving the MAGID pipeline's efficiency. It assures the generated images satisfy user-set standards in three domains: \textbf{Image-Text Matching}, \textbf{Image Quality}, and \textbf{Image Safety}.

\textbf{1- Image-text Matching}: We use the CLIP score \cite{radford2021learning} to validate the match between the image and the LLM model's utterance. A low CLIP score triggers image regeneration, with the count determined as a hyperparameter. In this work, we set the regeneration count to two. 

\textbf{2- Image Quality}: Images are rated based on an aesthetic score from \cite{schuhmann2022laion, author2023github}, which uses CLIP embedding followed by an MLP. This model identifies artifacts in the diffusion model outputs. A threshold of 0.51 efficiently detects most artifacts, prompting image regeneration for scores below this.

\textbf{3- Image Safety}: Image safety, particularly against NSFW content, is crucial. While many models assess this, few unsafe images were found in our dataset, indicating our process's reliability.

This robust QA ensures that MAGID can output relevant, high-quality, and safe images.

\subsubsection{Feedback Loop}

Should the diffusion model produce an image that does not meet the quality assurance module's stipulations, the issues might stem from the LLM model's prompt. Faulty prompts can yield low image-text matches or unsafe images. To mitigate this, our design, showcased in Figure \ref{fig:magi_f}, includes a feedback loop, instructing the LLM model to generate a better image description given regenerated images with previous image description continuously fall short of quality assurance standards. 

Figure \ref{fig:all_compare} displays a comparison of MAGID samples with two other datasets, MMDD \cite{lee2021constructing} and PhotoChat \cite{zang2021photochat}. A qualitative analysis shows MAGID yields quality comparable to real datasets, such as PhotoChat, and surpasses synthetic datasets like MMDD in generating high-quality multi-modal dataset. More examples are included in supplementary (section \ref{more_examples}).

\begin{table*}
\centering
\caption{Scanner module performance as measured by turn selection for image augmentation (accuracy, precision, recall, F1) and the resulting images from the generated descriptions (CLIP, MM-relevance, aesthetic) on the MMDialog dataset as ground-truth. The quality assurance module is \textbf{enabled}. We compare various LLMs powering the scanner module using chain of thought prompting.}
\vspace*{-0.3cm}
\label{enable_QA}
\small
\begin{tabular}{p{2.0cm}|M{1.1cm} M{1.1cm} M{1.1cm} M{1.1cm} M{1.5cm} M{1.9cm} M{1.1cm} M{1.1cm} M{1.2cm}}
\toprule
Model & Accuracy & Precision & Recall & \!\!F1 score\!\! & \!\!CLIP score\!\! & \!\!MM-Relevance\!\! & Aesthetic & \#images\\
\midrule
\!\!GPT 4 & \textbf{67.24}\% & \textbf{70.49}\% & \textbf{46.87}\% & \textbf{0.56} & \textbf{0.27}& \textbf{294.52} & 0.57 & 1359 \\
\!\!GPT 3.5 & 63.54\% & 69.43\% & 33.97\% & 0.46 & 0.26& 293.51 & \textbf{0.58}  &  1001 \\
\!\!Falcon-40b-Ins.\!\!\!\! & 58.93\% & 61.26\% & 24.13\% & 0.35 & 0.25 & 254.50  & \textbf{0.58}  & 794\\
\!\!Koala 13b & 56.28\% & 62.33\% & 6.91\% & 0.12 & 0.25 & 243.31 & 0.57 & 223 \\
\!\!Llama 13b & 57.10\% & 60.00\% & 13.64\% & 0.22 & 0.25 & 247.99 & 0.57 & 460 \\
\!\!OpenLLaMA & 57.94\% & 64.36\% & 12.69\% & 0.21 & 0.25 & 250.96 & \textbf{0.58} & 390 \\
\!\!Vicuna 13b & 58.77\% & 66.60\% & 14.38\% & 0.24 & 0.26 & 255.18 & 0.57 & 506 \\
\!\!MMDialogue\footnote{Ground Truth} & N/A  & N/A & N/A & N/A & 0.262 & N/A & 0.47 & 2717 \\
\bottomrule
\end{tabular}
\vspace*{-0.3cm}
\label{table:qunt_eval_w_qa}
\end{table*}

\section{Evaluation}

We scrutinize the efficacy and applicability of the multi-modal dataset generated by MAGID. Here are three pivotal questions we addressed in evaluation:

\begin{enumerate}[leftmargin=12pt]

\item How does MAGID quantitatively compare against real multi-modal datasets? $\rhd$ Section \ref{sec:quantitative_eval} 
\vspace*{-0.25cm}
\item Can MAGID create a multi-modal dataset with human-eye perceptible quality like a real one? $\rhd$ Section \ref{sec:human_eval} 
\vspace*{-0.25cm}
\item What is the impact of scanner prompt tuning and the quality assurance module on MAGID? $\rhd$ Section \ref{sec:ablation}

\end{enumerate}

The first and third question delves into a quantitative analysis, probing the accuracy and quality of the data generated by MAGID. Moreover, the second question is crucial, as a failure of MAGID to meet human evaluation standards would result in a low-quality training dataset that is unable to get positive human-centric assessments.  

In addition, in supplementary (section \ref{Downstream}), we have studied training multimodal model with MAGID and compared it with using real images for training.

\subsection{Quantitative Evaluation}
\label{sec:quantitative_eval}

\paragraph{Setup.}
Addressing the first question, a multi-dimensional evaluation assessed the image quality and accuracy of MAGID in selecting right utterances. To fairly compare MAGID’s general-use applicability, we only utilized prompt engineering to guide the LLM model to select the right utterances. In this regard, as a ground truth, we selected human-human interaction datasets MMDialog and PhotoChat, and removed images from their test sets and employed MAGID to transform the text-only data into a multi-modal dataset.

For the LLM-based model, we adopted a range of models, including GPT-4 \cite{openai2023gpt}, GPT-3.5 \cite{openai2023gpt}, Falcon-40b-Instruct \cite{penedo2023refinedweb}, Koala 13b \cite{geng2023koala}, LLaMA 13b \cite{touvron2023llama}, OpenLLaMA \cite{touvron2023llama}, and Vicuna 13b \cite{vicuna2023}. For image generation, SDXL 1.0 was consistently utilized across all models. We present the results of the MMDialog dataset here, and the PhotoChat results are included in supplementary (section \ref{photochat_res_sup}). In these experiments, we have set the threshold for the CLIP model at 0.21 and the aesthetic score threshold of 0.51. We used grid search to find these hyper-parameters. More details on computational cost is provided in supplementary (section \ref{cost}).


\vspace{-0.5cm}
\begin{table*}
\centering
\caption{Human Evaluation results of MAGID created datasets versus a retrieval-based synthetic dataset, MMDD, and two real datasets, MMDialouge and PhotoChat, where the mean shows the percentage of time the dialogues in one dataset were preferred among participants. (Q1: more realistic dialogue? Q2: images in which dialogue provide more knowledge?, Q3: better text-image matched?, Q4: better context-image matched?, Q5: more engaging?, Q6: hegher image quality?) }
\vspace*{-0.25cm}
\label{tab:human_mmdd}
\small
\begin{tabular}{p{0.5cm}|M{1.23cm} M{1.23cm} M{1.23cm}|M{1.23cm} M{1.23cm} M{1.23cm}|M{1.23cm} M{1.23cm} M{1.23cm}}
\toprule
& \multicolumn{3}{|c}{(a) MAGID vs. MMDD} & \multicolumn{3}{|c}{(b) MAGID vs. MMDialogue} & \multicolumn{3}{|c}{(c) MAGID vs. PhotoChat} \\\midrule
\# & Mean MAGID & Mean MMDD & Gwet's AC1 & Mean MAGID & \!\!Mean MMDial.\!\! & Gwet's AC1 & Mean MAGID & Mean Photo. & Gwet's AC1 \\
\midrule
Q1 & \textbf{96.29}\% & 3.71\% & 0.74 
& 48.17\% & \textbf{51.83}\% & 0.63
& \textbf{58.11}\% & 41.89\% & 0.47  \\

Q2 & \textbf{96.29}\% & 3.71\% &  0.89 
& 49.33\% & \textbf{50.67}\% & 0.65 
& \textbf{68.24}\% & 31.76\% & 0.71\\

Q3 & \textbf{89.11}\% & 10.89\% & 0.75 
& \textbf{52.72}\% & 47.28\% & 0.54
& \textbf{64.90}\% & 35.10\% & 0.53 \\

Q4 & \textbf{91.11}\% & 8.89\% & 0.83 
& 46.31\% & \textbf{53.69}\% & 0.65
& \textbf{61.98}\% & 38.02\% & 0.54 \\

Q5 & \textbf{95.57}\% & 4.43\% & 0.89 
& \textbf{51.94}\% & 48.06\% & 0.63
&  \textbf{64.02}\% & 35.98\% & 0.61 \\

Q6 & \textbf{80.92}\% & 19.08\% & 0.65 
& \textbf{63.90}\% & 36.10\% & 0.55
& \textbf{69.99}\% & 30.01\% & 0.64  \\\bottomrule
\end{tabular}
\vspace*{-0.35cm}
\label{table:human_eval}
\end{table*}
\begin{table}[t!]
\centering
\caption{Utterance selection accuracy using three different prompts on MMDialogue (ground-truth), where ZS, FS, and CoT stand for zero-shot, few-shot, and chain of thought respectively.}
\vspace*{-0.2cm}
\label{tab:my_label}
\small
\begin{tabular}{p{0.9cm}|M{1.2cm} M{1.2cm} M{1.2cm} M{1.0cm}}
\toprule
\!\!Prompt & Accuracy &  Precision & Recall  & F1 score \\
\midrule
\!\!ZS \!\! & 65.53\% & 73.12\% & 36.16\% & 0.48 \\
\!\!FS \!\! & 63.89\% & 69.67\% & 34.45\% & 0.46\\
\!\!CoT \!\! & \textbf{68.51}\% & \textbf{73.37}\% & \textbf{47.32}\% & \textbf{0.57}\\
\bottomrule
\end{tabular}

\vspace*{-0.4cm}
\label{table:abl_prompt}
\end{table}

\vspace{0.5cm}
\paragraph{Result.}
Table \ref{table:qunt_eval_w_qa} presents the performance of various LLM models on the MMDialog dataset. The table quantifies MAGID's response generation using different LLM models in comparison to the MMDialog dataset. The first column lists the LLM models used, while the subsequent four columns measure accuracy, precision, recall, and F1 score in choosing the correct utterance to be augmented with an image. The CLIP score gauges image-text matching, and the MM-Relevance, as introduced in \cite{feng2022mmdialog}, denotes the similarity between responses. In our context, it determines the resemblance of the produced image to the MMDialog's original image. The next column, the aesthetic score, indicates the image quality as discussed in \cite{author2023github}. Last row presents the ground truth dataset, highlighting the CLIP score, image count, and aesthetic quality of its images.

From the table, it is evident that GPT-4 and GPT-3.5 outperforms other models across all metrics. Notably, the CLIP and aesthetic scores of MAGID using GPT-4 and GPT-3.5 surpass even the ground truth values. In the next section, we also examine image-text matching and image quality in our human evaluation for MAGI against other datasets to test if it is aligned with our quantitative findings.

\subsection{Human Evaluation}
\label{sec:human_eval}

\paragraph{Setup.}
We conducted a human evaluation using a website with questionnaire. Participants viewed two dialogues: one with an image from MAGID and another from datasets MMDD \cite{lee2021constructing}, PhotoChat \cite{zang2021photochat}, or MMDialog \cite{feng2022mmdialog}. MAGID used GPT-4 as its Language Model and SDXL 1.0 for image generation. From the mentioned datasets, we selected 20 dialogues each, totaling 60 dialogues, and replaced their images with MAGID's. During evaluation, participants compared MAGID's multi-modal dialogues with the originals, without information about the dialogue origins.

For each dialogue pair (one from MAGID and one from the benchmark datasets), participants responded to the following prompts:
\begin{enumerate}[start=1,label={Q\arabic*:}]
\item Which dialogue appears more realistic? 
\vspace*{-0.25cm}
\item Which dialogue's images convey greater knowledge?
\vspace*{-0.25cm}
\item In which dialogue is there better match between images and the immediately preceding text?
\vspace*{-0.25cm}
\item In which dialogue do the images more closely match with the overall conversation context?
\vspace*{-0.25cm}
\item Which dialogue is more engaging?
\vspace*{-0.25cm}
\item Which dialogue features higher quality images?
\end{enumerate}
Respondents selected from binary choices (Dialogue A or Dialogue B) for each prompt. 

For this evaluation, 15 human annotators provided their answers. Schema of the website interface are available in the Supplementary materials (section \ref{human_eval_sup}).

\begin{table*}
\vspace{-0.45cm}
\centering
\caption{Ablation results of the MAGID framework with and without the quality assurance (QA) module. Results on turn selection and image quality performance across four LLMs on MMDialog (ground-truth) are shown. The first four rows are the results with the QA module, while the last four are the results without. The system prompt is chain of thought.}
\vspace*{-0.3cm}
\small
\begin{tabular}{p{2.0cm} |M{1.1cm} M{1.1cm} M{1.1cm} M{1.1cm} M{1.5cm} M{1.9cm} M{1.1cm} M{1.1cm} M{1.2cm}}
\toprule
Model & Accuracy & Precision & Recall & \!\!F1 score\!\! & \!\!CLIP score\!\! & \!\!MM-Relevance\!\! & Aesthetic & \#images\\\midrule
\!\!GPT 4 & \textbf{67.24}\% & \textbf{70.49}\% & \textbf{46.87}\% & \textbf{0.56} & \textbf{0.27}& \textbf{294.52} & 0.57 & 1359 \\
\!\!GPT 3.5 & 63.54\% & 69.43\% & 33.97\% & 0.46 & 0.26& 293.51 & \textbf{0.58}  &  1001 \\
\!\!Falcon-40b-Ins.\!\!\!\! & 58.93\% & 61.26\% & 24.13\% & 0.35 & 0.25 & 254.50  & 0.58 & 794 \\
\!\!OpenLLaMA & 57.94\% & 64.36\% & 12.69\% & 0.21 & 0.25 & 250.96 & \textbf{0.58} & 390 \\\midrule
\end{tabular}
\begin{tabular}{p{2.0cm} |M{1.1cm} M{1.1cm} M{1.1cm} M{1.1cm} M{1.5cm} M{1.9cm} M{1.1cm} M{1.1cm} M{1.2cm}}
\toprule
\!\!GPT 4 & 67.86\% & 69.70\% & \textbf{50.64}\% &\textbf{0.59} & \textbf{0.27} & \textbf{282.25} & 0.55 & 1485 \\
\!\!GPT 3.5 & \textbf{68.51}\% & \textbf{73.37}\% & 47.32\% & 0.57 & 0.26 & 278.16 & 0.55 & 1109 \\
\!\!Falcon-40b-Ins.\!\!\!\! & 56.77\% & 53.58\% & 28.80\% & 0.37 & 0.23 & 224.59  & 0.55 & 1075\\
\!\!OpenLLaMA & 58.92\% & 62.50\% & 21.51\% & 0.32 & 0.21 & 213.56 & \textbf{0.56} & 696 \\
\bottomrule
\end{tabular}
\vspace*{-0.3cm}
\label{table:qunt_eval_wo_qa}
\end{table*}

\paragraph{Result.}
Table \ref{table:human_eval} displays MAGID's results against MMDD, MMDialog, and PhotoChat datasets. The `Mean MAGID' column shows the percentage of annotators favoring MAGID, while `Mean Other' indicates those preferring the alternative dataset. Gwet’s AC1 measure, found in the last column, was used to assess inter-annotator reliability. It offers stability over Cohen’s Kappa \cite{wongpakaran2013comparison} and is more resilient to outliers (For more explanation, please refer to Supplementary Materials section \ref{gwet}.). 

From Table \ref{table:human_eval}(a), it's evident that annotators favored MAGID over the synthetically generated MMDD dataset across all question categories. Moreover, the high Gwet’s AC1 value indicates a strong consensus among annotators in choosing MAGID over MMDD. In contrast, when examining Table \ref{table:human_eval}(b), annotators exhibited a slight preference for the authentic MMDialog dataset in terms of realism. Notably, the Gwet’s AC1 value is considerably lower here than in the MMDD results, suggesting a reduced consensus among annotators. Nevertheless, MAGID outperformed MMDialog in terms of image quality and image-text matching. \textbf{Such findings affirm our quantitative evaluations and showcase the potential of generative AI in producing superior data sources for training.} As for the PhotoChat dataset (Table \ref{table:human_eval}(c)), while it is constructed from authentic human interactions, human participants were told to mock real conversation. Interestingly, our annotators slightly leaned towards MAGID over PhotoChat. This outcome suggests MAGID's promising capability to serve as an alternative to Mechanical Turk in the development of multi-modal datasets.

\subsection{Ablation Study of MAGID}
\label{sec:ablation}

We conducted ablation studies on (1) using different prompts for utterance identification and (2) investigating the impact of our quality assurance (QA) module.

\subsubsection{Prompts for Scanner}
\label{prompt_exp}

Table \ref{table:abl_prompt} displays the outcomes of three prompt strategies, namely Zero-shot (ZS) prompting, Few-shot prompting (FS), and Chain of Thought (CoT) prompting, as applied to the GPT-3.5 model for MAGID. These results are reported for the MMDialog dataset, with quality assurance deactivated, to solely measure the accuracy of the LLM model. Notably, the Chain of Thought strategy outperforms the other two across all evaluated metrics.

\subsubsection{Impact of QA Module}

Table \ref{table:qunt_eval_wo_qa} showcases the performance of four LLM models in MAGID, contrasting when the QA module is either enabled or disabled. A perusal of Table \ref{table:qunt_eval_wo_qa} reveals a decline in the aesthetic score, MM-Relevance, and CLIP score across all models upon the deactivation of QA. Moreover, a noticeable decrement in the precision of most models is observable, validating that the QA module bolsters MAGID by enhancing precision in pinpointing the optimal utterance for image generation. In contrast, disabling QA leads to an elevation in recall, attributable to MAGID selecting a more extensive array of utterances for image generation, thereby reducing the ratio of false negatives. Future research could explore the development of a refined QA module capable of elevating the recall rate for the entire pipeline.

\vspace{-0.2cm}
\section{MAGID Dataset}
\label{dataset_poc}
\vspace{-0.2cm}
As a proof of concept, and consistent with studies like \cite{lee2021constructing}, we employed text-only datasets such as DailyDialog \cite{li2017dailydialog}, Persona-Chat \cite{zhang2018personalizing}, and PhotoChat \cite{zang2021photochat} (by replacing its images with MAGID) to generate a multi-modal dataset \footnote{The link to dataset: https://github.com/amazon-science/MAGID} of 53,620 dialogues. Based on the results of our experiments, we used GPT-3.5 to transform 47,868 input dialogues and GPT-4 to augment the rest. Table \ref{tab:magid_dataset} shows the statistics of the generated dataset with MAGID. The data and the code will be made available to the public upon acceptance.

\begin{table}[t!]
\centering
\caption{Statistics of the MAGID dataset.}
\vspace*{-0.3cm}
\label{tab:magid_dataset}
\small
\begin{tabular}{p{4cm}|M{1.2cm} M{1.2cm}}
\toprule
\!\! Category & Train & Test\\
\midrule
\!\!Total dialogues \!\! & 47643 & 5977  \\
\!\!Avg length of dialogues \!\! & 11.76 & 11.36 \\
\!\!Avg length of sentences \!\! & 9.77 & 9.60 \\
\!\!Total images \!\! & 67951 & 10229 \\
\bottomrule
\end{tabular}

\vspace*{-0.3cm}
\end{table}

\vspace{-0.4cm}
\section{Conclusion}
\vspace{-0.2cm}
We presented a generative, fully automated pipeline designed to transform text-only datasets into multi-modal variants, harnessing the power of LLMs through prompt engineering. This solution addresses limitations faced by preceding methods, notably in terms of data privacy, accessibility, constrained image distribution, and occurrences of unsuitable or non-consensual content. Crucially, our pipeline permits the substitution of real, potentially privacy-compromising images with synthetic counterparts. We thoroughly evaluated our multi-modal data generation method using human assessment, quantitative analyses with various LLMs, and an in-depth ablation study. The promising results highlight generative AI's capability to stand as an alternative to traditional data generation methods, like mechanical turk.

Looking ahead, our dataset paves the way for developing large multi-modal language models that can engage with users via both text and visuals.

\section*{Limitations}

This paper predominantly concentrates on augmenting the privacy, diversity, and quality of multi-modal dataset generation by employing LLM and diffusion models. Although utilizing generative diffusion models can mitigate issues related to privacy breaches—given these models are also trained on extensive volumes of web images—they are susceptible to copyright infringement \cite{aboutalebi2023deepfakeart}. Addressing this issue exceeds the ambit of this paper and presents a compelling avenue for future work.

Moreover, the current work exclusively emphasizes image and text modalities. Extending considerations to additional modalities—such as video sharing, voice sharing, and more—is recommended for subsequent research endeavors. In addition, fine-tunning of large language model to generate image is left to future works. 

Improving generated image consistency in the dialogue is another important aspect that can further improve the quality of the generated multi-modal dataset by MAGID. Employing more recent diffusion models such as DALL-E 3 \cite{betker2023improving} can address this problem as they can make more consistent image generation. 
In this regard, in the section \ref{limilim} of Supplementary materials, we have included further examples that shows the limitations of the proposed MAGID pipeline. 

In conclusion, the enhancement of our quality assurance module is pivotal for developing more realistic multi-modal datasets from text-only inputs. In this regard, works like \cite{tian2023stablerep} already showed that using synthesized images is effective. This work prioritizes aspects like aesthetic score, clip score, and safety. Future research can explore additional elements to further refine and add realism to the transformation into multi-modal outputs. 

{\small
\bibliography{main}
}

\newpage
\appendix
\label{sec:appendix}

\hspace{-1cm}{\LARGE\bfseries Supplementary}
\vspace{1cm}

\section{COT \& FS Prompts}
\label{COT_prompt}

In the paper, we referenced the \textbf{Few Shot} and \textbf{Chain of Thought} prompts, which can be found in Figures \ref{fig:few_shot} and \ref{fig:cot}, respectively. When generating multi-modal versions from each text-only input dataset, it became evident that distinct prompting is necessary for the chain of thoughts due to variations in the format of the input text. 

\section{PhotoChat results}
\label{photochat_res_sup}
As mentioned in section \ref{sec:quantitative_eval}, here we have included the results of different LLM on PhotoChat dataset. Table \ref{photochat_res} shows the results. Overall, GPT 3.5 shows better performance compared with other LLM models. As it can be seen, the precision is significantly lower compared with the results reported on MMDialogue dataset (Table \ref{enable_QA})  and that is because this dataset is limited to only one image per dialogue while our pipeline does not have such restriction. 

\begin{table*}[]
\caption{Different LLM model testing on PhotoChat (ground-truth). Quality Assurance module is enabled. The system prompt is chain of thoughts.}
\vspace*{-0.2cm}
\small
\begin{tabular}{p{2.0cm}|M{1.1cm} M{1.1cm} M{1.1cm} M{1.1cm} M{1.5cm} M{1.9cm} M{1.1cm} M{1.1cm} M{1.2cm}}
\toprule
Model & Accuracy & Precision & Recall & \!\!F1 score\!\! & \!\!CLIP score\!\! & \!\!MM-Relevance\!\! & \#images & Aesthetic\\
\midrule
\!\!GPT 3.5 & 86.11\% & \textbf{28.62}\% & 25.91\% & 0.27 & 0.25 & 313.64 & 87 & 0.57 \\
\!\!Falcon-40b-Ins.\!\!\!\! & 88.10\% & 28.04\% & 11.83\% & 0.17 & 0.24 & 303.68  & 403 & \textbf{0.58}\\
\!\!Koala 13b & 89.61\% & 30.43\% & 2.94\% & 0.05 & 0.24 & 283.44 & 92 & 0.61 \\
\!\!Llama 13b & 87.32\% & 20.79\% & 9.54\% & 0.13 & 0.23 & 244.36 & 433 & 0.59 \\
\!\!OpenLLaMA & 88.75\% & 27.31\% & 8.03\% & 0.12 & 0.23 & 270.36 & 696 & 0.59 \\
\!\!Vicuna 13b & 88.40\% & 25.48\% & 8.35\% & 0.13 & 0.24 & 244.97 & 602 & 0.55 \\
\!\!PhotoChat & N/A  & N/A & N/A & N/A & 0.213 & N/A & 961 & 0.49 \\
\bottomrule
\end{tabular}
\label{photochat_res}
\end{table*}

\begin{figure*}[] 
    \centering
    \begin{tcolorbox}[colback={promptbgcolor},title={\small Few-shot example prompt},enhanced,attach boxed title to top right={yshift=-3.5mm, xshift=-5mm}, colbacktitle=white, coltitle=black, top=12pt]  
        \small
    \begin{alltt}
    \textcolor{specialtokencolor}{- query: >} 
        Utterance: 0: So yeah, it was a mostly dismal year. But what's the best news you've
        read/saw/heard in 2016? (Anything from the personal to world affairs.) 
        Utterance: 1: grew 14 pumpkins on the formidable  strength of my chickens We are 
        all proud! Here's one 
        Utterance: 2: Very impressive! 
      \textcolor{specialtokencolor}{answer: >}
        \textcolor{slotcolor}{<result>} Utterance: 1: 14 pumpkins\textcolor{slotcolor}{</result>} 
    \textcolor{specialtokencolor}{- query: >}
        Utterance: 0: Working from home with a tie today! Plenty of Zoom in my life today!
        Utterance: 1: I keep a polo handy that I throw on and off for zoom calls. 
        Way to be extra fancy
      \textcolor{specialtokencolor}{answer: >}
        \textcolor{slotcolor}{<result>}Utterance: 0: Working from home with a tie\textcolor{slotcolor}{</result>}
    \end{alltt}
   \end{tcolorbox}
   \caption{The few-shot example prompt not only provides the format for both input and expected output along with a problem description but also includes multiple exemplars to elucidate the desired response from the LLM. Here only exemplars are included.}
   \label{fig:few_shot}
\end{figure*}

\begin{figure*}[] 
    \centering
    \begin{tcolorbox}[colback={promptbgcolor},title={\small Chain of Thoughts  prompt},enhanced,attach boxed title to top right={yshift=-3.5mm, xshift=-5mm}, colbacktitle=white, coltitle=black, top=12pt]  
        \small
    \begin{alltt}
    \textcolor{specialtokencolor}{- query: >} 
        Utterance: 0: So yeah, it was a mostly dismal year. But what's the best news you've
        read/saw/heard in 2016? (Anything from the personal to world affairs.) 
        Utterance: 1: grew 14 pumpkins on the formidable  strength of my chickens We are 
        all proud! Here's one 
        Utterance: 2: Very impressive! 
      \textcolor{specialtokencolor}{answer: >}
      \textcolor{slotcolor}{<reason>} Utterance 0 is just a descrption of last year without any information that 
        can be translated with image. We add photographic style as it is a personal sharing.
        Utterance 1 on the other hand is talking about growing 14 pumpkins. This can be
        represented with image.\textcolor{slotcolor}{</reason>}
        \textcolor{slotcolor}{<result>} Utterance: 1: 14 pumpkins\textcolor{slotcolor}{</result>} 
    \textcolor{specialtokencolor}{- query: >}
        Utterance: 0: My attire for the SA Hip Hop Awards
        Utterance: 1: Are you a supporter of Kaizer Chiefs?...lol. Gorgeous!!
      \textcolor{specialtokencolor}{answer: >}
        \textcolor{slotcolor}{<reason>}In Utterance 0 contains the sentence "My outfit for 
        the SA hip hop awards" which shows
        the person is willing to share her outfit\textcolor{slotcolor}{</reason>}
        \textcolor{slotcolor}{<result>}Utterance: 0: My outfit for the SA hip hop awards \textcolor{slotcolor}{</result>}
    \end{alltt}
   \end{tcolorbox}
   \caption{The chain of thoughts prompt, building upon the system prompt provided in the few-shot example prompt, also incorporates the detailed reasoning on utterance selection.}
   \label{fig:cot}
\end{figure*}

\section{Image Generator Ablation Study}
\label{diffuse_compare}
Table \ref{table:abl_stable} shows the performance of different diffusion models \cite{podell2023sdxl, rombach2022high}. The results are taken from MMDialog dataset and the quality assurance module is disabled to report the results without filtering unwanted ones. It is clear that SDXL 1.0 and SDXL 0.9 has very similar performance and higher aethetic score compared with Stable Diffusion 2.0. All models have similar CLIP score which is predictable as they are given the same prompt for image generation. 

\begin{table}[b!]
\centering
\caption{Testing different Stable diffusion models with MAGID pipeline }
\vspace*{-0.3cm}
\small
\begin{tabular}{p{3cm}|M{1.3cm} M{1.2cm} M{1.2cm}}
\toprule
\!\! Model  & Aesthetic Score & CLIP Score\\
\midrule
\!\!SDXL 1.0 \!\!  & 0.56 & 0.26 \\
\!\!SDXL 0.9 \!\!  & \textbf{0.57} & 0.26 \\
\!\!Stable Diffusion 2.0 \!\!  & 0.53 & 0.26 \\
\bottomrule
\end{tabular}

\vspace*{-0.1cm}
\label{table:abl_stable}
\end{table}

\section{Human evaluation}
\label{human_eval_sup}

\begin{figure*}[htbp]
\includegraphics[width=1.0\textwidth]{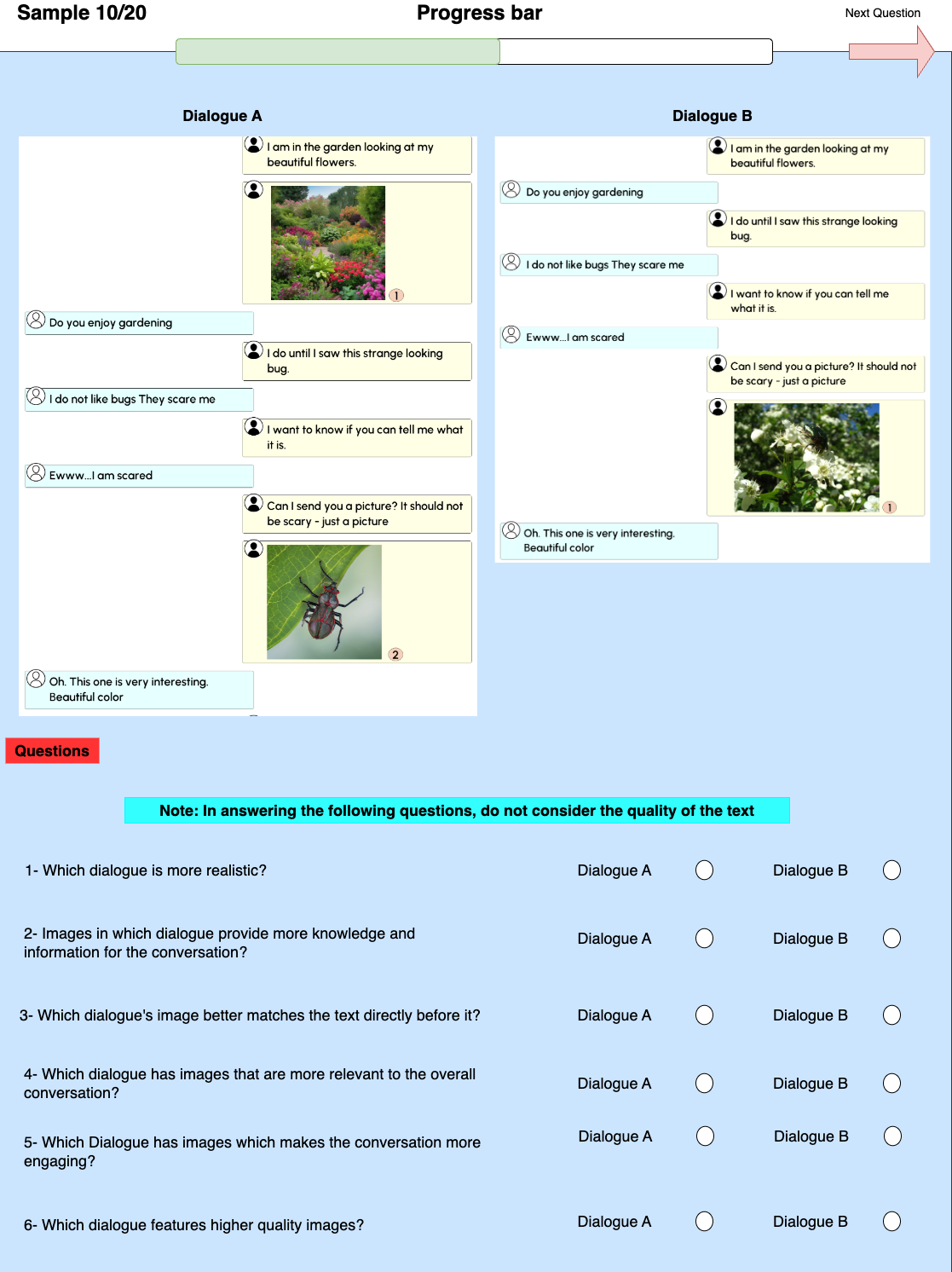}

\caption{Schema of the website used to perform human evaluation. }
\label{fig:website}
\end{figure*}

To collect answers from annotators, we created a website with a schema shown in Figure~\ref{fig:website}. For each question, annotators were given two screenshots of the same dialogue, one generated by MAGID and the other from a source dataset (PhotoChat, MMDialog, or MMDD). At the start of the annotation session, annotators were instructed to ignore the conversation text and focus only on the images and image-text matching. Fifteen annotators completed the task, each making 20 comparisons.

\section{Downstream Training}
\label{Downstream}

Here, we study how much MAGID can impact training a multi-modal model when changing the original image with synthetic one generated by MAGID. In addition, we also compare it with benchmark cases when no image is present in the training and with MMDD \cite{lee2021constructing} approach to include image in the dialogue. 
In this regard, we used the same architecture suggested in \cite{lee2023building} which is visionTextDualEncoder from Huggingface \cite{wolf2019huggingface} which projects the encoding of image with the the embedding of text to a shared common space. For encoding of image we used ViT \cite{dosovitskiy2020image}, and for processing the text we used pretrained DialoGPT \cite{zhang2019dialogpt}.
While the input is multi-modal, the output is text only. In this task, we omit the last text utterance and the model should predict it given the prior image and text.

We fine-tuned the model on MMDialog dataset and the results are reported in Table \ref{table:down}. For this experiment, we used the learning rate of $0.00005$ with Adam Optimizer. In Table \ref{table:down}, we show the results on the test set when training set images is coming from MMDialogue, MAGID, MMDD and the case where the images are omitted. For MMDD, we used the same code they used to inject image into text-only dialogue to make the comparison possible. For this expeiment, the training set consists of 5156 dialogues and the test set consists of 633 dialogues sampled from MMDialogue dataset. 

\begin{table}[ht!]
\centering
\caption{Downstream training. The model used is DialoGPT + ViT. BLUE score is in percentage. }
\vspace*{-0.3cm}
\small
\begin{tabular}{p{2cm}|M{1cm} M{1cm} M{1cm} M{1cm} M{1cm} M{1cm}}
\toprule
\!\! Dataset  & PPL & BLEU-1 & BLEU-2 & distinct-1 & distinct-2\\
\midrule
\!\!MMDialogue \!\!  & 73.09  & \textbf{8.3} & \textbf{3.9} & 0.94 & 0.965\\
\!\!MAGID \!\!  & \textbf{70.99} & 7.9 & \textbf{3.9} & 0.94 & \textbf{0.971} \\
\!\!MMDD \!\!  & 78.86 & 7.5 & 3.0 & 0.93 & 0.963 \\
\!\!No image \!\!  & 78.88 & 7.9 & 3.0 & 0.92 & 0.952 \\
\bottomrule
\end{tabular}

\vspace*{-0.1cm}
\label{table:down}
\end{table}


As it can be seen, when we use the source image as training set (MMDialog), we achieve highest BLEU score \cite{papineni2002bleu}. The perplexity of the model using MAGID is lowest which shows the model is more confident in making the prediction. In addition, the distinct score \cite{liu2022rethinking} which shows the diversity of response is highest with MAGID which can be attributed to higher image-text match provided with MAGID images. It is important to note that since MMDialog dataset is a real dataset, the quality of images shared does not necessarily matches the text and this can make the model less confident and results in higher perplexity. On the other hand, the images generated by MAGID is more controlled. 

For this experiment we used 4 NVIDIA RTX GPU each with 24 GiB memory and the training took for a full day. 

\section{Experiment Computational Cost}
\label{cost}

For running MAGID pipeline, it can be run with one GPU with NVIDIA RTX with 24 GiB memory. 

\section{Discussion on Inter-rater Reliability Measure Choice}
\label{gwet}

In Section 4.2, we employed Gwet's AC1 for evaluating the consistency among reviewers, opting not to use Cohen's Kappa due to its susceptibility to outliers and potential for showing inconsistent results despite high average scores across all participants. As detailed in the study by Wongpakaran et al. (2013), Gwet's AC1 is recognized for its greater consistency in inter-rater reliability assessments when compared to Cohen’s Kappa, alongside its enhanced resilience to outliers, providing a more reliable measure for our analysis \cite{wongpakaran2013comparison}. This approach ensures a more stable and accurate assessment of reviewer consistency, mitigating the impact of anomalies on the reliability scores.

\section{Further examples of MAGID}
\label{more_examples}

\begin{figure*}[htbp]
\includegraphics[width=1.0\textwidth]{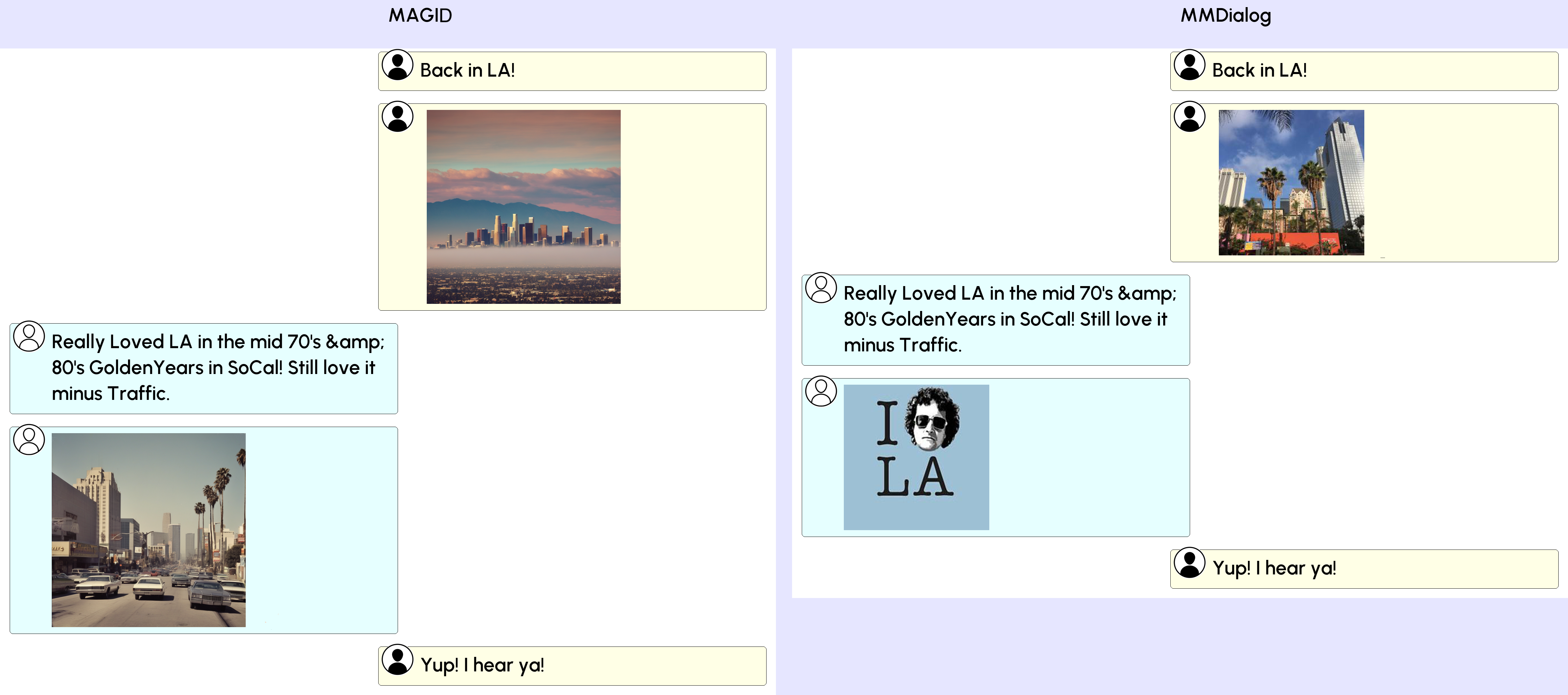}

\caption{MAGID (left) versus MMDialog (right) }
\label{fig:more2}
\end{figure*}

\begin{figure*}[htbp]
\includegraphics[width=1.0\textwidth]{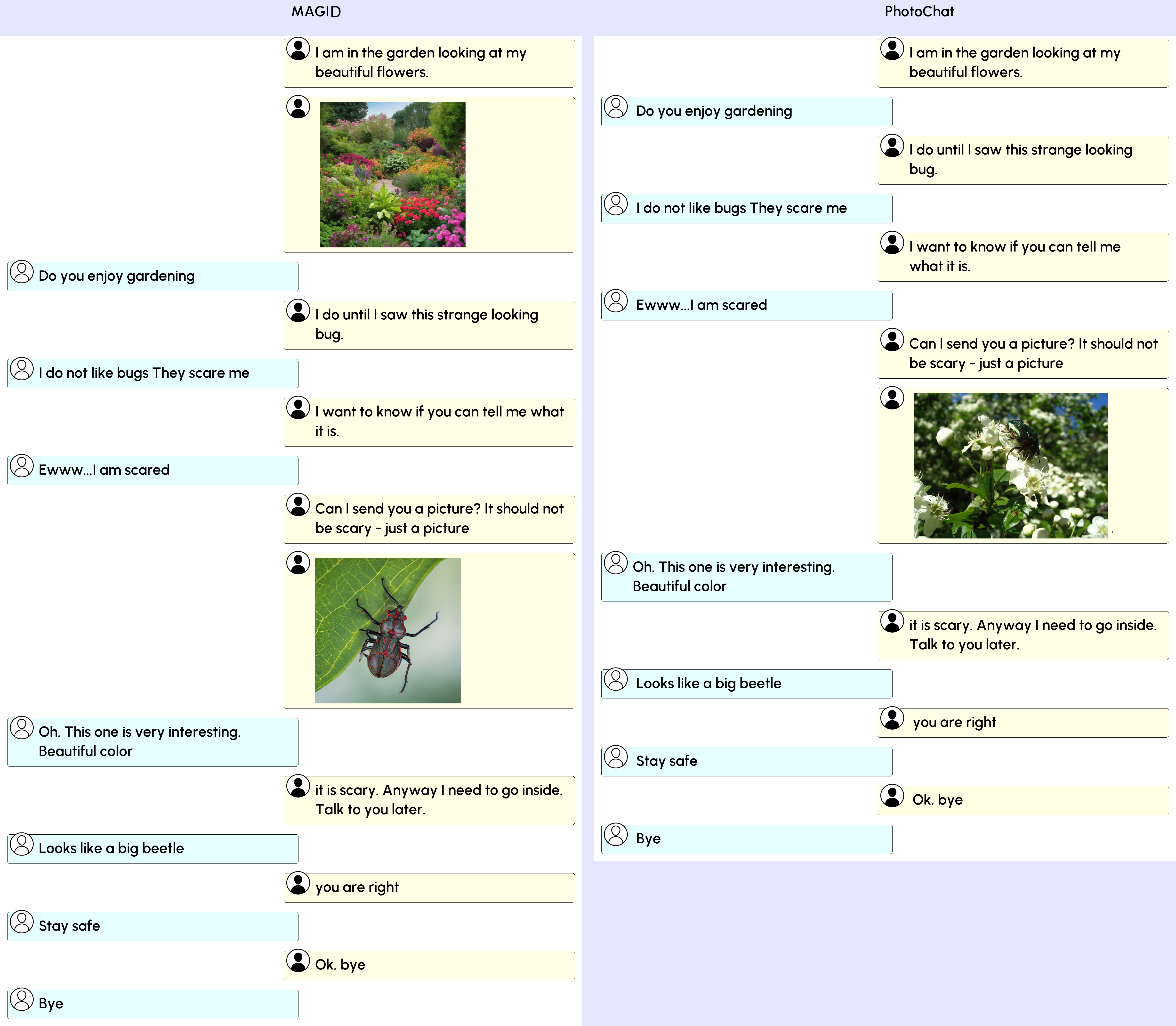}

\caption{MAGID (left) versus PhotoChat (right)}
\label{fig:more1}
\end{figure*}

\begin{figure*}[htbp]
\includegraphics[width=1.0\textwidth]{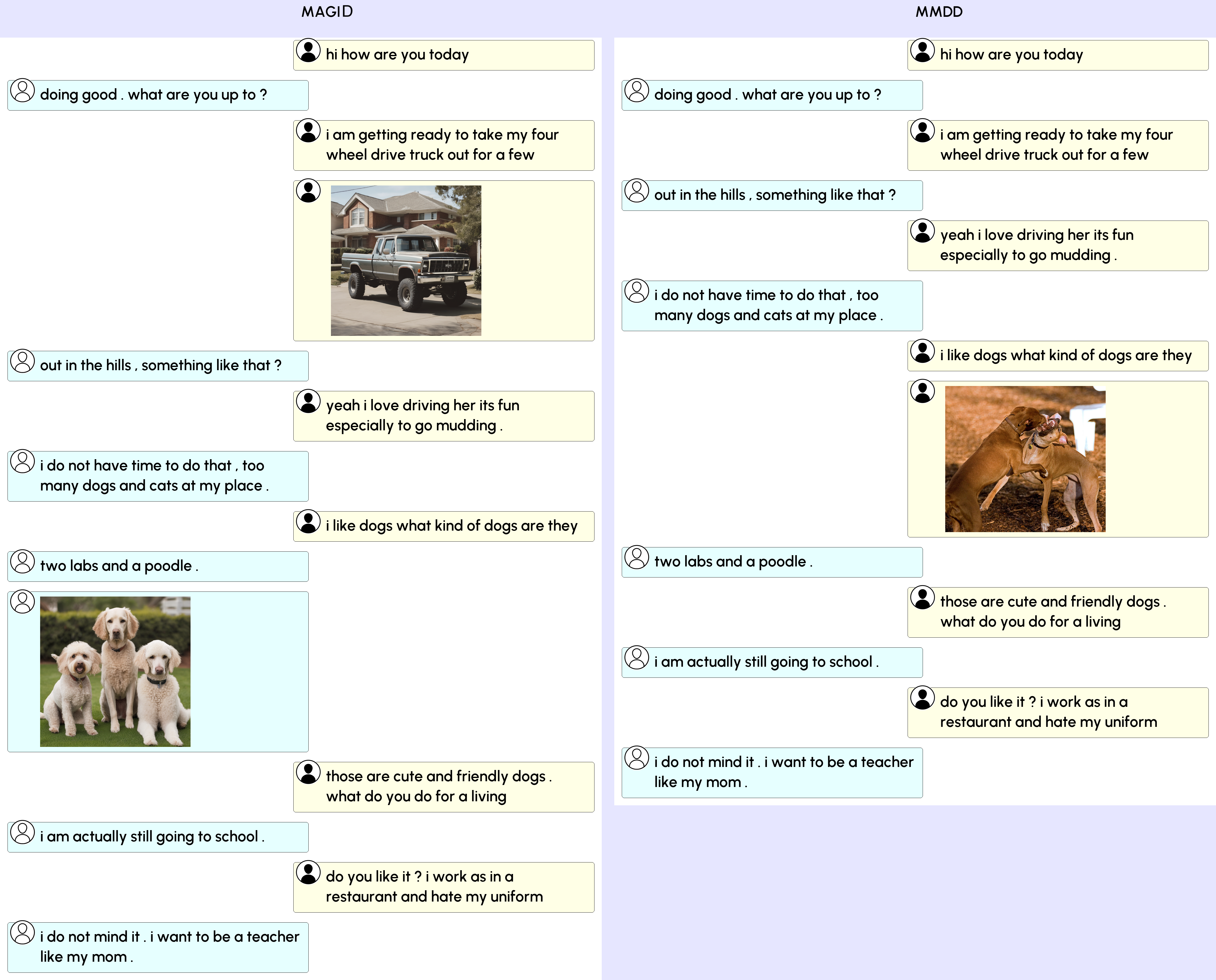}

\caption{MAGID (left) versus MMDD (right)}
\label{fig:more3}
\vspace*{-0.4cm}
\end{figure*}

Figures \ref{fig:more2}, \ref{fig:more1}, and \ref{fig:more3} provide more examples on comparing MAGID with MMDialog, PhotoChat, and MMD. 

\section{Experiment Setting}

For determining the threshold for image-text matching and aesthetic score, we employed cross-validation on the validation set. In this regard, the threshold for CLIP score was set for 0.21 and the threshold for the aesthetic score was set for 0.51. Based on our observations, we established a protocol where a generated image could fail up to two times before being discarded and triggering the feedback loop. This approach ensured a balance between generating high-quality images and maintaining efficient processing. In all our experiments, we used SDXL 1.0 model for image generation. 

\section{Limitations}  \label{limilim}

\begin{figure}[htbp]
\centering
\includegraphics[width=\linewidth]{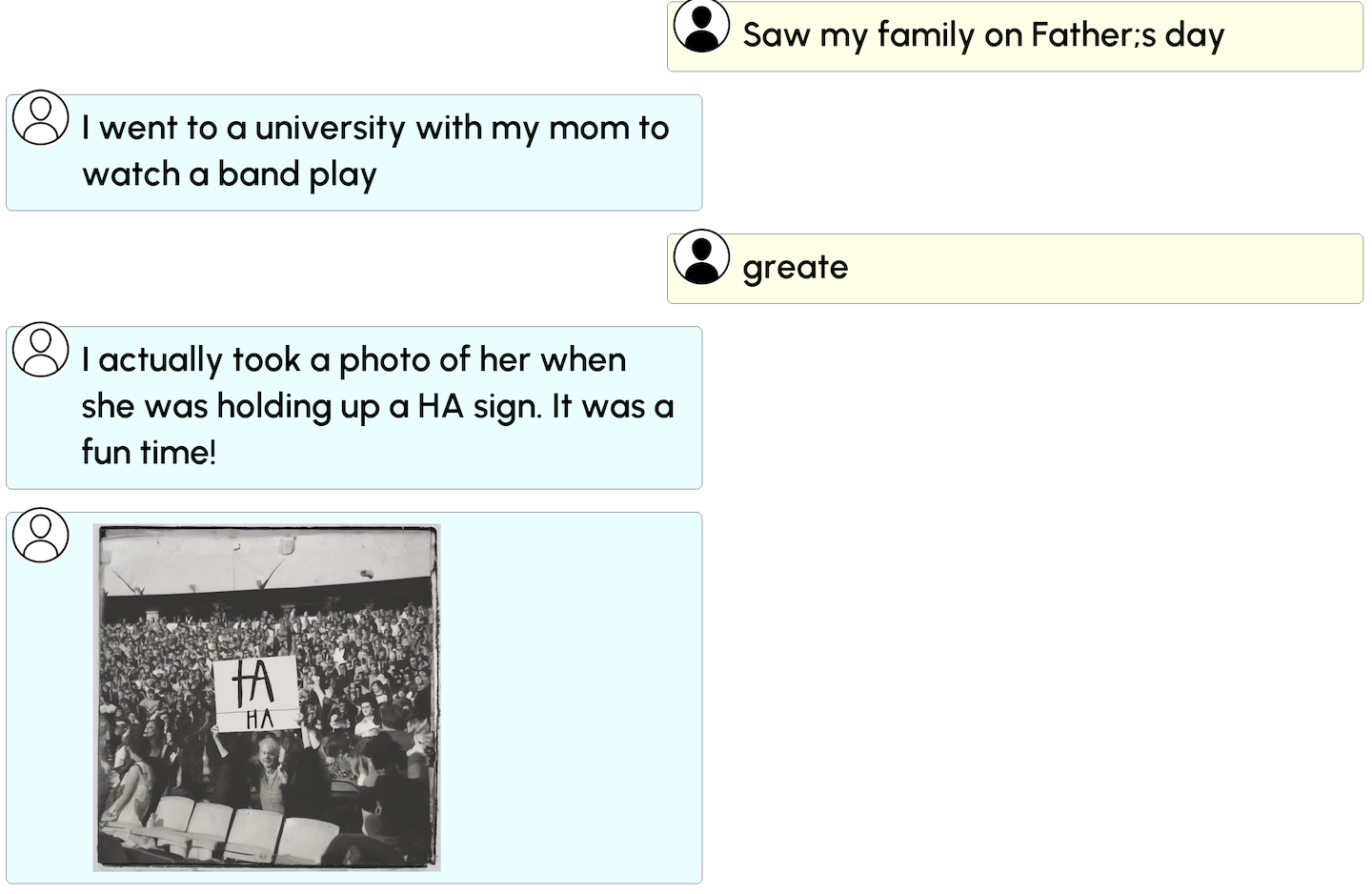}
\caption{Generated image by MAGID fails to properly show the sign HA}
\label{fig:lim1}
\vspace{2cm}
\end{figure}

\begin{figure}[htbp]
\centering
\includegraphics[width=\linewidth]{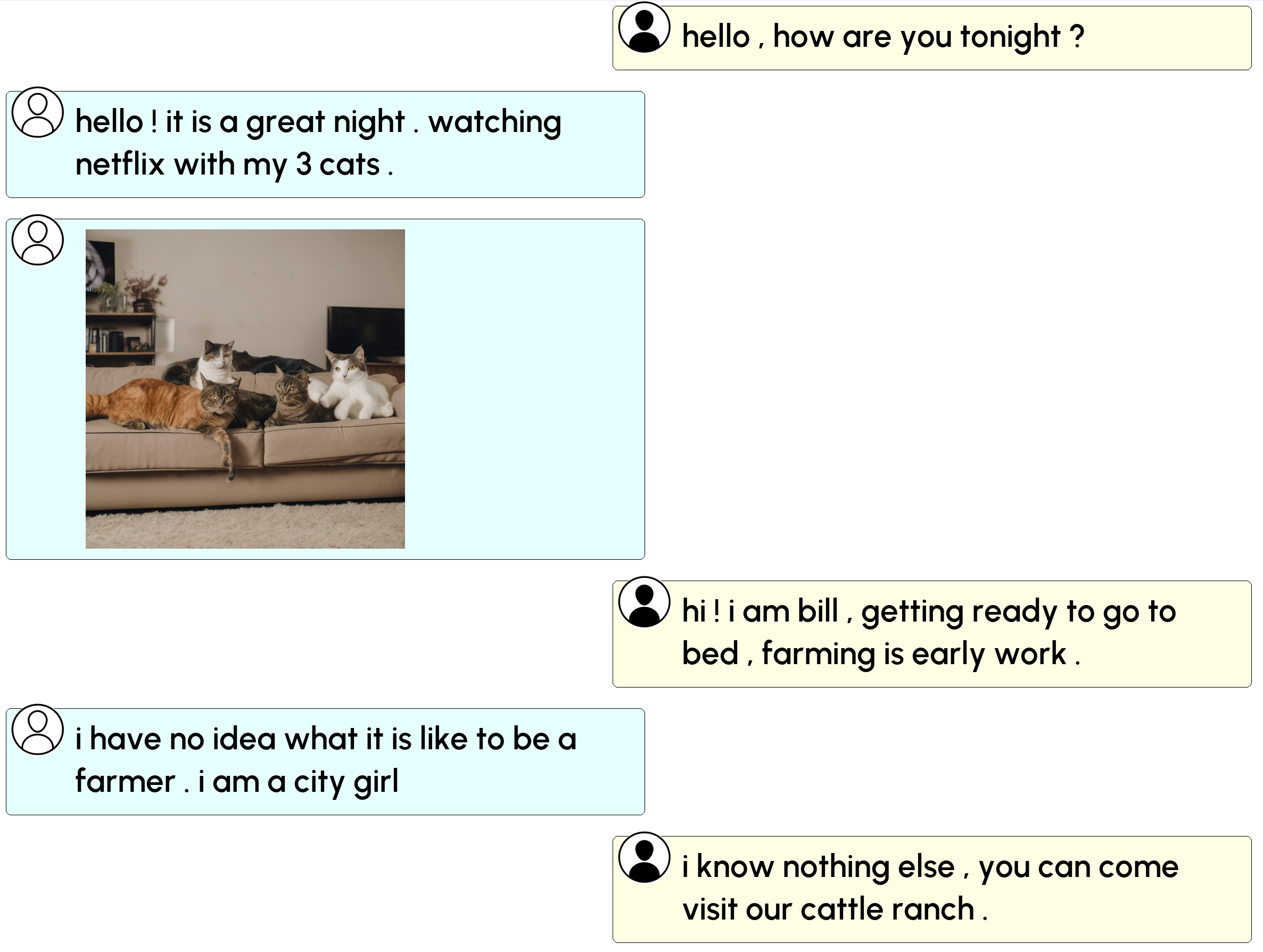}
\caption{Generated image by MAGID fails to properly shows 4 cats instead of 3}
\label{fig:lim2}
\end{figure}

\begin{figure}[t]
\centering
\vspace{-17cm}
\includegraphics[width=\linewidth]{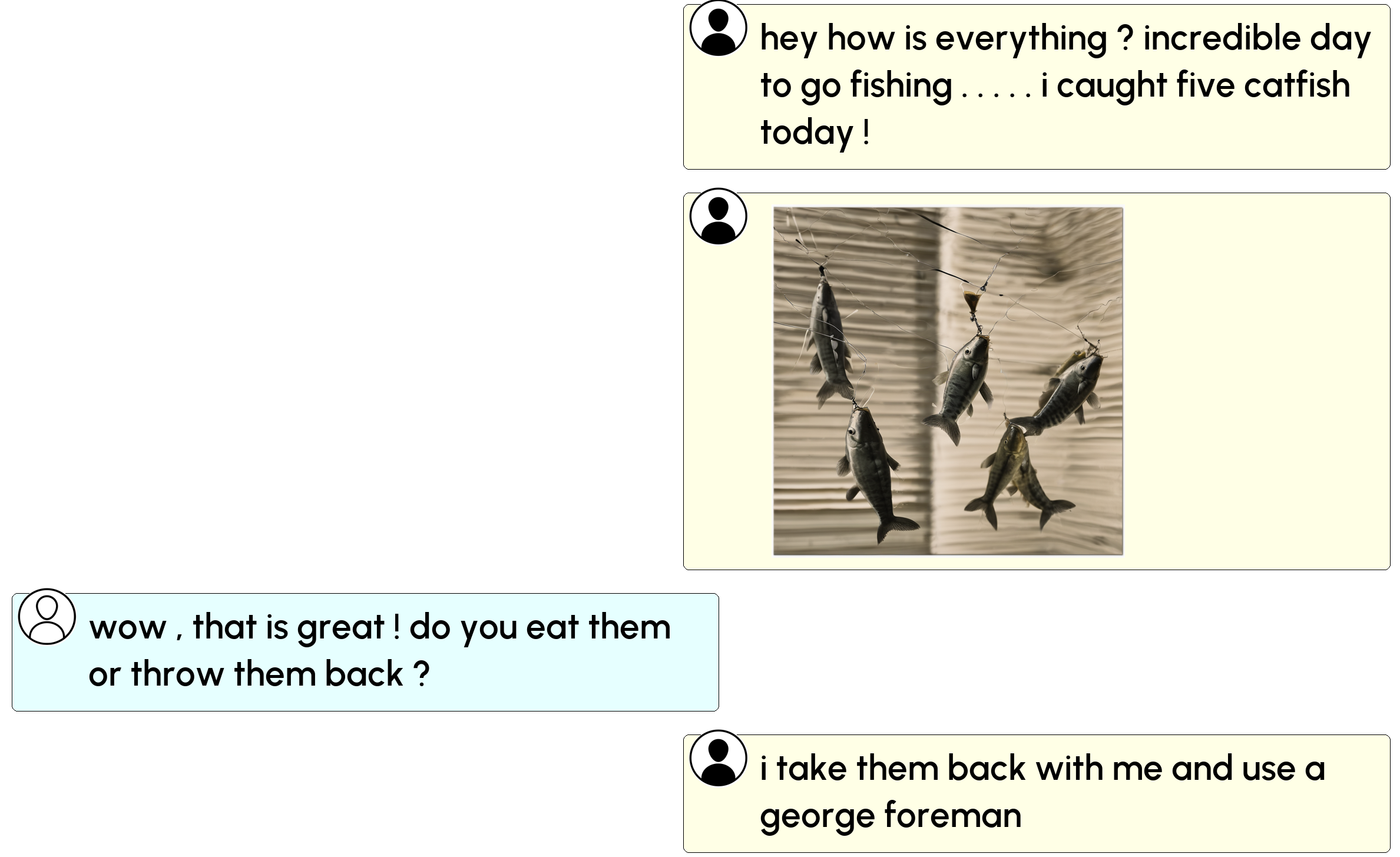}
\caption{Generated image by MAGID fails to properly shows 6 fishes instead of 5}
\label{fig:lim3}
\end{figure}

In Figures \ref{fig:lim1}, \ref{fig:lim2}, and \ref{fig:lim3}, we showcase the most common scenarios were MAGID can fail to generate the image which properly supports the preceding utterance. Specifically, figure \ref{fig:lim1} shows a common example, where the generated image usually fails to put the proper text sign in the generated image. In Figures \ref{fig:lim2} and \ref{fig:lim3} showcase the examples where the generated image does not follow the correct description in terms of number object should exist in the image. We believe using more advanced diffusion models like DALL-E 3 should mitigate this problem. 

\end{document}